%% file: private_GP.tex
\documentclass{article}

\usepackage{times}
\usepackage{graphicx} 
\usepackage{subfigure}

\usepackage{natbib}

\usepackage{algorithm}
\usepackage{algorithmic}
\usepackage{amssymb}
\usepackage{amsmath}
\usepackage{amsthm}
\usepackage{nicefrac}

\usepackage{hyperref}

\newcommand{\x}{\mathbf{x}}
\newcommand{\xbar}{\overline{\mathbf{x}}}
\newcommand{\ybar}{\overline{y}}
\newcommand{\w}{\mathbf{w}}

\newcommand{\I}{\mathbf{I}}

\newcommand{\K}{\mathbf{K}}

\newcommand{\Ac}{\mathcal{A}}
\newcommand{\X}{\mathbf{X}}
\newcommand{\Xc}{\mathcal{X}}
\newcommand{\blam}{\boldsymbol{\lambda}}
\newcommand{\Tc}{\mathcal{T}}
\newcommand{\Vc}{\mathcal{V}}
\newcommand{\vb}{\mathbf{v}}
\newcommand{\GP}{\mathcal{G}\mathcal{P}}
\newcommand{\Prob}{\Pr}
\newcommand{\Rc}{\mathcal{R}}
\newcommand{\Rbb}{\mathbb{R}}
\newcommand{\BO}{{\mbox{\scriptsize{BO}}}}

\DeclareMathOperator*{\argmax}{arg\,max}
\DeclareMathOperator*{\argmin}{arg\,min}
\DeclareMathOperator{\Lap}{Lap}

\usepackage{color}

\newtheorem{assumption}{Assumption}
\newtheorem{define}{Definition}
\newtheorem{thm}{Theorem}

\newtheorem{coro}{Corollary}

\usepackage[accepted]{icml2015_2}

\icmltitlerunning{Differentially Private Bayesian Optimization}

\begin{document}

%

%

\twocolumn[
\icmltitle{Differentially Private Bayesian Optimization}

\icmlauthor{Matt J. Kusner}{mkusner@wustl.edu}
\icmlauthor{Jacob R. Gardner}{gardner.jake@wustl.edu}
\icmlauthor{Roman Garnett}{garnett@wustl.edu}
\icmlauthor{Kilian Q. Weinberger}{kilian@wustl.edu}
\icmladdress{Computer Science \& Engineering,
			Washington University in St. Louis}

\icmlkeywords{boring formatting information, machine learning, ICML}

\vskip 0.3in
]


\begin{abstract}
Bayesian optimization is a powerful tool for fine-tuning the hyper-parameters of a wide variety of machine learning models. The success of machine learning has led practitioners in diverse real-world settings to learn classifiers for practical problems. As machine learning becomes commonplace, Bayesian optimization becomes an attractive method for practitioners to automate the process of classifier hyper-parameter tuning. A key observation is that the data used for tuning models in these settings is often sensitive. Certain data such as genetic predisposition, personal email statistics, and car accident history, if not properly private, may be at risk of being inferred from Bayesian optimization outputs. To address this, we introduce methods for releasing the best hyper-parameters and classifier accuracy privately.  Leveraging the strong theoretical guarantees of differential privacy and known Bayesian optimization convergence bounds, we prove that under a GP assumption these private quantities are also near-optimal. Finally, even if this assumption is not satisfied, we can use different smoothness guarantees to protect privacy.
\end{abstract}

\section{Introduction}
\input{intro.tex}

\section{Background}
\input{background.tex}


\input{noise.tex}

\section{Without observation noise}
\label{sec:exact}
\input{exact.tex}




We prove Corollary \ref{thm:best_f_noise} and Theorem \ref{thm:f_utility} in the supplementary material. 
We have demonstrated that in the noisy and noise-free settings we can release private near-optimal hyper-parameter settings $\tilde{\lambda}$ and function evaluations $\tilde{v}, \tilde{f}$. However, the analysis thus far assumes the hyper-parameter set is finite: $| \Lambda | \!<\! \infty$. It is possible to relax this assumption, using an analysis similar to \citep{srinivas2009gaussian}. We leave this analysis to the supplementary material.

\section{Without the GP assumption}
\label{sec:no_gp}
\input{no_gp.tex}

\section{Results}
\input{results.tex}


\section{Related work}
\input{related.tex}

\section{Conclusion}
\input{conclusion.tex}

\bibliography{bibliography}
\bibliographystyle{icml2015}

\end{document}

%% file: intro.tex
Machine learning is increasingly used in application areas with sensitive data. For example, hospitals use machine learning to predict if a patient is likely to be readmitted soon~\citep{yu2013predicting}, webmail providers classify spam emails from non-spam~\citep{weinberger2009feature}, and insurance providers forecast the extent of bodily injury in car crashes~\citep{chong2005traffic}.

In these scenarios data cannot be shared legally, but companies and hospitals may want to share 
hyper-parameters and validation accuracies through publications or other means. However, data-holders must be careful, as even a small amount of information can compromise privacy.

Which hyper-parameter setting yields the highest accuracy can reveal sensitive information about individuals in the validation or training data set, reminiscent of reconstruction attacks described by \citet{dwork2013algorithmic} and \citet{dinur2003revealing}.
For example, imagine updated hyper-parameters are released right after a prominent public figure is admitted to a hospital. 
If a hyper-parameter is known to correlate strongly with a particular disease the patient is suspected to have, an attacker could make a direct correlation between the hyper-parameter value and the individual. 

To prevent this sort of attack, we develop a set of algorithms that automatically fine-tune the hyper-parameters of a machine learning algorithm 
while provably preserving differential privacy~\citep{dwork2006calibrating}. 
Our approach leverages recent results on Bayesian optimization~\citep{snoek2012practical,hutter2011sequential,bergstra2012random,gardner2014bayesian}, training a Gaussian process (GP)~\citep{rasmussen2006gaussian} to accurately predict and maximize the validation gain of hyper-parameter settings. 
We show that the GP model in Bayesian optimization allows us to release noisy final hyper-parameter settings to protect against aforementioned privacy attacks, while only sacrificing a tiny, bounded amount of validation gain.



Our privacy guarantees hold for releasing the best hyper-parameters and best validation gain. 
Specifically our contributions are as follows:
\begin{itemize}
\item We derive, to the best of our knowledge, the first framework for Bayesian optimization with provable differential privacy guarantees,
\item We develop variations both with and without observation noise, and 
\item We show that even if our validation gain is not drawn from a Gaussian process, we can guarantee differential privacy under different smoothness assumptions.
\end{itemize}

We begin with background on Bayesian optimization and differential privacy we will use to prove our guarantees.

%% file: background.tex
In general, our aim will be to protect the privacy of a validation dataset of sensitive records $\Vc \subseteq \Xc$ (where $\Xc$ is the collection of all possible records) when the results of Bayesian optimization depends on $\Vc$.


\paragraph{Bayesian optimization.}
Our goal is to maximize an unknown function $f_{\Vc}\colon \Lambda \rightarrow \mathbb{R}$ that depends on some validation dataset ${\Vc} \subseteq \Xc$:
\begin{equation}
\max_{\lambda \in \Lambda} f_{\Vc}(\lambda). \label{eq:bayesopt}
\end{equation}
It is important to point out that all of our results hold for the general setting of eq.~(\ref{eq:bayesopt}), but throughout the paper, we use the vocabulary of a common application: that of machine learning hyper-parameter tuning. In this case $f_{\Vc}(\lambda)$ is the gain of a learning algorithm evaluated on validation dataset $\Vc$ that was trained with hyper-parameters $\lambda \in \Lambda \subseteq \Rbb^d$.  

As evaluating $f_{\Vc}$ is expensive (e.g., each evaluation requires training a learning algorithm), Bayesian optimization gives a procedure for selecting a small number of locations to sample $f_{\Vc}$: $[\lambda_1, \ldots, \lambda_T] \!=\! \blam_{T} \!\in\! \Rbb^{d\times T}$. 
Specifically, given a current sample $\lambda_t$, we observe a validation gain $v_t$ such that $v_t = f_{\Vc}(\lambda_t) + \alpha_t$, where $\alpha_t \sim {\cal N}(0,\sigma^2)$ is Gaussian noise with possibly non-zero variance $\sigma^2$. Then, given $v_t$ and previously observed values $v_1, \ldots, v_{t-1}$, Bayesian optimization updates its belief of $f_{\Vc}$ and samples a new hyper-parameter $\lambda_{t+1}$. Each step of the optimization proceeds in this way.

To decide which hyper-parameter to sample next, Bayesian optimization places a prior distribution over $f_{\Vc}$ and updates it after every (possibly noisy) function observation. 
One popular prior distribution over functions is the Gaussian process
${\cal G}{\cal P}\bigl(\mu(\cdot),k(\cdot,\cdot)\bigr)$ \citep{rasmussen2006gaussian}, parameterized by a mean function $\mu(\cdot)$ (we set $\mu = 0$, w.l.o.g.) and a kernel covariance function $k(\cdot,\cdot)$. 
Functions drawn from a Gaussian process have the property that any finite set of values of the function are normally distributed. Additionally, given samples $\blam_{T} = [\lambda_1, \ldots, \lambda_T]$ and observations $\vb_T = [v_1, \ldots, v_T]$,
the GP posterior mean and variance has a closed form: 
\begin{align}
\mu_T(\lambda) =& \; k(\lambda,\blam_T)( \K_T + \sigma^2 \I  )^{-1} \vb_T \nonumber \\
k_T(\lambda,\lambda') =&\; k(\lambda,\lambda') - k(\lambda,\blam_T)( \K_T + \sigma^2 \I  )^{-1} k(\blam_T,\lambda') \nonumber \\
\sigma^2_T(\lambda) =& \; k_T(\lambda,\lambda), \label{eq:post_GP}
\end{align}
where $k(\lambda,\blam_T) \!\in\! \Rbb^{1 \times T}$ is evaluated element-wise on each of the $T$ columns of $\blam_T$. As well, $\K_T = k(\X_T,\X_T) \!\in\! \Rbb^{T \times T}$ and $\lambda \in \Lambda$ is any hyper-parameter. As more samples are observed, the posterior mean function $\mu_T(\lambda)$ approaches $f_{\Vc}(\lambda)$. 

One well-known method to select hyper-parameters $\lambda$ maximizes the \emph{upper-confidence bound} (UCB) of the posterior GP model of $f_{\Vc}$ \citep{auer2002finite,srinivas2009gaussian}:
\begin{align}
\lambda_{t+1} \triangleq \argmax_{\lambda \in \Lambda} \mu_t(\lambda) + \sqrt{\beta_{t+1}}\sigma_t(\lambda), \label{eq:ucb}
\end{align}
where $\beta_{T+1}$ is a parameter that trades off the \emph{exploitation} of maximizing $\mu_t(\lambda)$ and the \emph{exploration} of maximizing $\sigma_t(\lambda)$. \citet{srinivas2009gaussian} proved that given certain assumptions on $f_{\Vc}$ and fixed, non-zero observation noise $(\sigma^2 \!>\! 0)$, selecting hyper-parameters $\lambda$ to maximize eq.~(\ref{eq:ucb}) is a no-regret Bayesian optimization procedure: $\lim_{T \to \infty} \frac{1}{T} \sum_{t = 1}^T f_{\Vc}(\lambda^*) - f_{\Vc}(\lambda_t) = 0$, where $f_{\Vc}(\lambda^*)$ is the maximizer of eq.~(\ref{eq:bayesopt}). For the no-noise setting, \citet{de2012exponential} give a UCB-based no-regret algorithm.

\paragraph{Contributions.}
Alongside maximizing $f_{\Vc}$, we would like to guarantee that if $f_{\Vc}$ depends on (sensitive) validation data, we can release information about $f_{\Vc}$ so that the data ${\Vc}$ remains private. Specifically, we may wish to release (a) our best guess $\hat{\lambda} \triangleq \argmax_{t \leq T} f_{\Vc}(\lambda_t)$ of the true (unknown) maximizer $\lambda^*$ and (b) our best guess $f_{\Vc}(\hat{\lambda})$ of the true (also unknown) maximum objective $f_{\Vc}(\lambda^*)$. 
The primary question this work aims to answer is: how can we release private versions of $\hat{\lambda}$ and $f_{\Vc}(\hat{\lambda})$ 
that are close to their true values, or better, the values $\lambda^*$ and $f_{\Vc}(\lambda^*)$? We give two answers to these questions. The first will make a Gaussian process assumption on $f_{\Vc}$, which we describe immediately below. The second, described in Section \ref{sec:no_gp}, will utilize Lipschitz and convexity assumptions to guarantee privacy in the event the GP assumption does not hold.

\paragraph{Setting.}
For our first answer to this question, let us define a Gaussian process over hyper-parameters $\lambda,\lambda' \in \Lambda$ \emph{and} datasets $\Vc, \Vc' \!\subseteq\! \Xc$ as follows: $\GP\bigl(0, k_1(\Vc,\Vc') \otimes k_2(\lambda,\lambda')\bigr)$. A prior of this form is known as a multi-task Gaussian process \citep{bonilla2008multi}. Many choices for $k_1$ and $k_2$ are possible. The function $k_1(\Vc,\Vc')$ defines a set kernel (e.g., a function of the number of records that differ between $\Vc$ and $\Vc'$). For $k_2$, we focus on either the squared exponential: $k_2(\lambda,\lambda') = \exp\bigl(-\| \lambda - \lambda' \|_2^2 / (2 \ell^2)\bigr)$ or Mat\'{e}rn kernels: (e.g., for $\nu = 5/2$, $k_2(\lambda,\lambda') = (1 + \sqrt{5} r / \ell + (5 r^2) / (3\ell^2) ) \exp(- \sqrt{5} r /\ell )$, for $r = \| \lambda - \lambda' \|_2$), for a  fixed $\ell$,
as they have known bounds on the maximum information gain \citep{srinivas2009gaussian}. Note that as defined, the kernel $k_2$ is normalized (i.e., $k_2(\lambda,\lambda) = 1$).

\begin{assumption}
\label{assume:gp}
We have a problem of type (\ref{eq:bayesopt}), where all possible dataset functions $[f_1, \ldots, f_{2^{|\Xc|}}]$ are GP distributed $\GP\bigl(0, k_1(\Vc,\Vc') \otimes k_2(\lambda,\lambda')\bigr)$ for known kernels $k_1, k_2$, for all $\Vc,\Vc' \!\subseteq\! \Xc$ and $\lambda,\lambda' \!\in\! \Lambda$, where $|\Lambda| \!\leq\! \infty$.
\end{assumption}

Similar Gaussian process assumptions have been made in previous work \citep{srinivas2009gaussian}.
For a result in the no-noise observation setting, we will make use of the assumptions of \citet{de2012exponential} for our privacy guarantees, as described in Section \ref{sec:exact}.

\subsection{Differential Privacy}
One of the most widely accepted frameworks for private data release is \emph{differential privacy} \citep{dwork2006calibrating}, which has been shown to be robust to a variety of privacy attacks \citep{ganta2008composition,sweeney1997weaving,narayanan2008robust}. Given an algorithm $\Ac$ that outputs a value $\lambda$ when run on dataset $\Vc$, the goal of differential privacy is to `hide' the effect of a small change in $\Vc$ on the output of $\Ac$. Equivalently, an attacker should not be able to tell if a private record was swapped in $\Vc$ just by looking at the output of $\Ac$. If two datasets $\Vc,\Vc'$ differ by swapping a single element, we will refer to them as \emph{neighboring} datasets. Note that any non-trivial algorithm (i.e., an algorithm $\Ac$ that outputs different values on $\Vc$ and $\Vc'$ for some pair $\Vc,\Vc' \subseteq \Xc$) must include some amount of randomness to guarantee such a change in $\Vc$ is unobservable in the output $\lambda$ of $\Ac$ \citep{dwork2013algorithmic}. The level of privacy we wish to guarantee decides the amount of randomness we need to add to $\lambda$ (better privacy requires increased randomness). Formally, the definition of differential privacy is stated below.
\begin{define}
\label{def:dp}
A randomized algorithm $\Ac$ is $(\epsilon,\delta)$-\textbf{differentially private} for $\epsilon,\delta \geq 0$ if for all $\lambda \in \emph{\mbox{Range}}(\Ac)$ and for all neighboring datasets $\Vc,\Vc'$ (i.e., such that $\Vc$ and $\Vc'$ differ by swapping one record) we have that
\begin{align}
\Prob\bigl[\Ac(\Vc) = \lambda\bigr] \leq e^{\epsilon} \Prob\bigl[ \Ac(\Vc') = \lambda\bigr] + \delta. \label{eq:dp}
\end{align}
\end{define}

The parameters $\epsilon, \delta$ guarantee how private $\Ac$ is; the smaller, the more private. The maximum privacy is $\epsilon = \delta = 0$ in which case eq.~(\ref{eq:dp}) holds with equality. This can be seen by the fact that $\Vc$ and $\Vc'$ can be swapped in the definition, and thus the inequality holds in both directions. If $\delta=0$, we say the algorithm is simply $\epsilon$-differentially private. For a survey on differential privacy we refer the interested reader to \citet{dwork2013algorithmic}.

There are two popular methods for making an algorithm $\epsilon$-differentially private: (a) the Laplace mechanism \citep{dwork2006calibrating}, in which we add random noise to $\lambda$ and (b) the exponential mechanism \citep{mcsherry2007mechanism}, which draws a random output $\tilde{\lambda}$ such that $\tilde{\lambda} \approx \lambda$. For each mechanism we must define an intermediate quantity called the \emph{global sensitivity} describing how much $\Ac$ changes when $\Vc$ changes.

\begin{define}
(Laplace mechanism) The \textbf{global sensitivity} of an algorithm $\Ac$ over all neighboring datasets $\Vc,\Vc'$ (i.e., $\Vc,\Vc'$ differ by swapping one record) is
\begin{align}
\Delta_{\Ac} \triangleq \max_{\Vc,\Vc' \subseteq \Xc } \| \Ac(\Vc) - \Ac(\Vc') \|_1. \nonumber
\end{align}
(Exponential mechanism) The \textbf{global sensitivity} of a function $q\colon \!\Xc \!\times\! \Lambda \!\rightarrow\! \mathbb{R}$ over all neighboring datasets $\Vc,\Vc'$ is
\begin{equation}
\Delta_q \triangleq \max_{\substack{\Vc,\Vc' \subseteq \Xc \\ \lambda \in \Lambda}} \| q(\Vc,\lambda) - q(\Vc',\lambda) \|_1. \nonumber
\end{equation}
\end{define}

%

The Laplace mechanism hides the output of $\Ac$ by perturbing its output with some amount of random noise.

\begin{define}
Given a dataset $\Vc$ and an algorithm $\Ac$, the \textbf{Laplace mechanism} returns $\Ac(\Vc) + \omega$, where $\omega$ is a noise variable drawn from $\Lap(\Delta_{\Ac} / \epsilon)$, the Laplace distribution with scale parameter $\Delta_{\Ac} / \epsilon$ (and location parameter $0$).
\end{define}

The exponential mechanism draws a slightly different $\tilde{\lambda}$ that is `close' to $\lambda$, the output of $\Ac$.


\begin{define}
Given a dataset $\Vc$ and an algorithm $\Ac(\Vc) \!=\! \argmax_{\lambda \in \Lambda} q(\Vc,\lambda)$, the \textbf{exponential mechanism} returns $\tilde{\lambda}$, where $\tilde{\lambda}$ is drawn from the distribution $\frac{1}{Z} \exp\bigl( \epsilon q(\Vc,\lambda) / (2 \Delta_q )\bigr)$, and $Z$ is a normalizing constant.
\end{define}

Given $\Lambda$, a possible set of hyper-parameters, we derive methods for privately releasing the best hyper-parameters and the best function values $f_{\Vc}$, approximately solving eq.~(\ref{eq:bayesopt}). 
We first address the setting with observation noise $(\sigma^2 \!>\! 0)$ in eq.~(\ref{eq:post_GP}) and then describe small modifications for the no-noise setting. For each setting we use the UCB sampling technique in eq.~(\ref{eq:ucb}) to derive our private results.



%% file: noise.tex
\section{With observation noise}
\label{sec:noise}
In general cases of Bayesian optimization, observation noise occurs in a variety of real-world modeling settings such as sensor measurement prediction \citep{krause2008near}. In hyper-parameter tuning, noise in the validation gain may be as a result of noisy validation or training features. 

\begin{algorithm}                      
\caption{ Private Bayesian Opt. (noisy observations)}          
\label{alg1}                           
\begin{algorithmic}                    
	\STATE \textbf{Input:} $\Vc$; $\Lambda \subseteq \mathbb{R}^d$; $T$; $(\epsilon,\delta)$; $\sigma^2_{\Vc,0}$; $\gamma_T$
	\STATE $\mu_{\Vc,0} = 0$
	\FOR{ $t = 1 \ldots T$ }
		\STATE $\beta_{t} \!=\! 2\log( |\Lambda| t^2 \pi^2 / (3 \delta) )$
		\STATE $\lambda_{t} \triangleq \argmax_{\lambda \in \Lambda} \mu_{\Vc, t-1}(\lambda) + \sqrt{\beta_{t}}\sigma_{\Vc, t-1}(\lambda)$
		\STATE Observe validation gain $v_{\Vc,t}$, given $\lambda_t$
		\STATE Update $\mu_{\Vc, t}$ and $\sigma^2_{\Vc, t}$ according to \eqref{eq:post_GP}
	\ENDFOR
	\STATE $c \!=\! 2 \sqrt{\bigl(1 \!-\! k(\Vc,\Vc')\bigr) \log \big(3 |\Lambda|/\delta\big) }$
	\STATE $q \!=\!  \sigma \sqrt{4 \log(3/\delta)}$
	\STATE $C_1 = 8/\log(1 + \sigma^{-2})$
	\STATE  Draw $\tilde{\lambda} \in \Lambda$ w.p. $\Prob[\lambda] \propto \mbox{exp}\Big(\frac{\epsilon \mu_{\Vc,T}(\lambda)}{2(2 \sqrt{\beta_{T+1}} + c)}\Big)$
	\STATE $v^* \!=\! \max_{t \leq T} v_{\Vc,t}$
	\STATE Draw $\theta\sim\! \Lap\!\Big[ \frac{\sqrt{C_1 \beta_T \gamma_T}}{\epsilon\sqrt{T}} + \frac{c}{\epsilon} + \frac{q}{\epsilon}   \Big]$
	\STATE $\tilde{v}=v^*+\theta$
	\STATE \textbf{Return:} $\tilde{\lambda},\tilde{v}$
\end{algorithmic}
\end{algorithm}


In the sections that follow, although the quantities $f, \mu, \sigma, v$ all depend on the validation dataset $\Vc$, for notational simplicity we will occasionally omit the subscript $\Vc$. Similarly, for $\Vc'$ we will often write: $f', \mu', {\sigma'}^{2}, v'$.


\subsection{Private near-maximum hyper-parameters}
\label{sec:x_noise}
In this section we guarantee that releasing $\tilde{\lambda}$ in Algorithm \ref{alg1} is private (Theorem \ref{thm:1}) and that it is near-optimal (Theorem \ref{thm:2}). Our proof strategy is as follows: we will first demonstrate the global sensitivity of $\mu_T(\lambda)$ with probability at least $1\!-\!\delta$. Then we will show show that releasing $\tilde{\lambda}$ via the exponential mechanism is $(\epsilon,\delta)$-differentially private. Finally, we prove that $\mu_T(\tilde{\lambda})$ is close to $f(\lambda^*)$, the true maximizer of eq.~(\ref{eq:bayesopt}).

\paragraph{Global sensitivity.}
As a first step we bound the global sensitivity of $\mu_T(\lambda)$ as follows:

\begin{thm}
\label{lem:sens_x_noise}
Given Assumption \ref{assume:gp}, for any two neighboring datasets $\Vc,\Vc'$ and for all $\lambda \!\in\! \Lambda$ with probability at least $1- \delta$ there is an upper bound on the global sensitivity (in the exponential mechanism sense) of $\mu_T$:
\begin{equation}
| \mu'_T(\lambda) - \mu_T(\lambda) | \leq  2\sqrt{\beta_{T+1}} + \sigma_1\sqrt{2\log\left(3 |\Lambda| / \delta\right)}, \nonumber
\end{equation}
for $\sigma_1\!=\!\sqrt{2\bigl(1\!-\! k_1(\Vc,\Vc')\bigr)}$, $\beta_{t} \!=\! 2\log\Bigl( \! |\Lambda| t^2 \pi^2 /(3 \delta) \! \Bigr)$.
\end{thm}

\emph{Proof.} Note that, by applying the triangle inequality twice, for all $\lambda \in \Lambda$,
\begin{align}
&| \mu'_T(\lambda) - \mu_T(\lambda) | \leq  \; | \mu'_T(\lambda) - f'(\lambda) | + | f'(\lambda) - \mu_T(\lambda) | \nonumber\\
&\leq  | \mu'_T(\lambda) - f'(\lambda) | + | f'(\lambda) - f(\lambda) | + |f(\lambda)-\mu_T(\lambda)|.\nonumber
\end{align}
We can now bound each one of the terms in the summation on the right hand side (RHS) with probability at least $\frac{\delta}{3}$. 
According to \citet{srinivas2009gaussian},  Lemma 5.1, we obtain $| \mu'_T(\lambda) - f'(\lambda) |\!\leq\! \sqrt{\beta_{T+1}} \sigma'_T(\lambda)$. The same can be applied to $| f(\lambda)-\mu_T(\lambda) |$. 
As $\sigma'_T(\lambda)\!\leq\! 1$, because $k(\lambda,\lambda)\!=\!1$, we can upper bound both terms by $2\sqrt{\beta_{T+1}}$.
 In order to bound the remaining (middle) term on the RHS recall that for a random variable $Z \!\sim\! {\cal N}(0,1)$ we have: $\Prob\bigl[|Z| \!>\! \gamma\bigr]\! \leq\! e^{-\gamma^2/2}$. For variables $Z_1,\dots Z_n\!\sim\! {\cal N}(0,1)$, we have, by the union bound, that $\Prob\bigl[\forall i,\ |Z_i|\!\leq \!\gamma\bigr]\geq 1-ne^{-\gamma^2/2} \triangleq 1-\frac{\delta}{3}$. If we set $Z=\frac{|f(\lambda) - f'(\lambda)|}{\sigma_1}$ and  $n\!=\!|\Lambda|$, we obtain  $\gamma \!=\!  \sqrt{2\log\bigl( 3 | \Lambda | / \delta\bigr)}$, 
which completes the proof.  \hfill$\blacksquare$

We remark that all of the quantities in Theorem \ref{lem:sens_x_noise} are either given or selected by the modeler (e.g, $\delta, T$). Given this upper bound we can apply the exponential mechanism to release $\tilde{\lambda}$ privately, as per Definition~\ref{def:dp}:


\begin{coro}
\label{thm:1}
Let $\Ac(\Vc)$ denote Algorithm \ref{alg1} applied on dataset $\Vc$. Given Assumption \ref{assume:gp}, $\tilde{\lambda}$ is $(\epsilon,\delta)$-differentially private, i.e.,
$\Prob\bigl[ \Ac(\Vc) \!=\! \tilde{\lambda}\bigr] \!\leq\! e^\epsilon \Prob\bigl[ \Ac(\Vc') \!=\! \tilde{\lambda} \bigr] \!+\! \delta$,  
for any pair of neighboring datasets $\Vc, \Vc'$.
\end{coro}

We leave the proof of Corollary \ref{thm:1} to the supplementary material. Even though we must release a noisy hyper-parameter setting $\tilde{\lambda}$, it is in fact near-optimal.

\begin{thm}
\label{thm:2}
Given Assumption \ref{assume:gp} the following near-optimal approximation guarantee for releasing $\tilde{\lambda}$ holds:
\begin{align}
\textstyle
\mu_T(\tilde{\lambda}) \geq f(\lambda^*) - 2\sqrt{\beta_{T}} - q - \frac{2 \Delta}{\epsilon}( \log |\Lambda| + a ) \nonumber
\end{align}
w.p.\ $\geq 1 - (\delta + e^{-a})$, where $\Delta =  2\sqrt{\beta_{T+1}} + c$  (for $\beta_{T+1}$, $c$, and $q$ defined as in Algorithm \ref{alg1}).
\end{thm}

\emph{Proof.}
In general, the exponential mechanism selects $\tilde{\lambda}$ that is close to the maximum $\lambda$ \citep{mcsherry2007mechanism}:
\begin{equation}
\textstyle
\mu_T(\tilde{\lambda}) \geq \max_{\lambda \in \Lambda} \mu_T(\lambda) - \frac{2 \Delta}{\epsilon}( \log |\Lambda| + a ), \label{eq:exp_utility}
\end{equation}
with probability at least $1 - e^{-a}$. Recall we assume that at each optimization step we observe noisy gain $v_t = f(\lambda_t) + \alpha_t$, where $\alpha_t \sim {\cal N}(0, \sigma^2)$ (with fixed noise variance $\sigma^2 \!>\! 0$). As such, we can lower bound the term $\max_{\lambda \in \Lambda} \mu_T(\lambda)$:
\begin{align}
\max_{\lambda \in \Lambda} \mu_T(\lambda) \geq& \; \underbrace{f(\lambda_T) + \alpha_T }_{v_T} \nonumber \\
f(\lambda^*) - \max_{\lambda \in \Lambda} \mu_T(\lambda) \leq& \; f(\lambda^*) - f(\lambda_T) + \alpha_T \nonumber \\
f(\lambda^*) - \max_{\lambda \in \Lambda} \mu_T(\lambda) \leq& \; 2\sqrt{\beta_T}\sigma_{T-1}(\lambda_T) + \alpha_T \nonumber \\
\max_{\lambda \in \Lambda} \mu_T(\lambda) \geq& \; f(\lambda^*) - 2\sqrt{\beta_T} + \alpha_T, \label{eq:up_max}
\end{align}
where the third line follows from \citet{srinivas2009gaussian}: Lemma 5.2 and the fourth line from the fact that $\sigma_{T-1}(\lambda_T) \leq 1$. 

As in the proof of Theorem \ref{lem:sens_x_noise}, given a normal random variable $Z \!\sim\! {\cal N}(0,1)$ we have: $\Prob[ |Z| \!\leq\! \gamma] \!\geq\! 1 \!-\! e^{-\gamma^2/2} := 1 \!-\! \frac{\delta}{2}$. Therefore if we set $Z \!=\! \frac{\alpha_T}{\sigma}$ we have $\gamma \!=\! \sqrt{2 \log(2/\delta)}$. This implies that $|\alpha_T| \!\leq\! \sigma \sqrt{2 \log(2/\delta)} \!\leq\! \sqrt{4 \log(3/\delta)} \!=\! q$ (as defined in Algorithm \ref{alg1}) with probability at least $1 \!-\! \frac{\delta}{2}$. Thus, we can lower bound $\alpha_T$ by $-q$. We can then lower bound $\max_{\lambda \in \Lambda} \mu_T(\lambda)$ in eq.~(\ref{eq:exp_utility}) with the right hand side of eq.~(\ref{eq:up_max}). Therefore, given the $\beta_T$ in Algorithm \ref{alg1}, \citet{srinivas2009gaussian}, Lemma 5.2 holds with probability at least $1 - \frac{\delta}{2}$ and the theorem statement follows. \hfill$\blacksquare$



\subsection{Private near-maximum validation gain}
\label{sec:f_noise}
In this section we demonstrate releasing the validation gain $\tilde{v}$ in Algorithm \ref{alg1} is private (Theorem \ref{lem:sense_max_y}) and that the noise we add to ensure privacy is bounded with high probability (Theorem \ref{thm:max_f_bound}). As in the previous section our approach will be to first derive the global sensitivity of the maximum $v$ found by Algorithm \ref{alg1}. Then we show releasing $\tilde{v}$ is $(\epsilon,\delta)$-differentially private via the Laplace mechanism. Perhaps surprisingly, we also show that $\tilde{v}$ is close to $f(\lambda^*)$.


\paragraph{Global sensitivity.}
We bound the global sensitivity of the maximum $v$ found with Bayesian optimization and UCB:

\begin{thm}
\label{lem:sense_max_y}
Given Assumption \ref{assume:gp}, and neighboring $\Vc,\Vc'$, we have the following global sensitivity bound (in the Laplace mechanism sense) for the maximum $v$, w.p. $\geq\! 1\!-\!\delta$
\begin{align}
| \max_{t \leq T} v'_t - \max_{t \leq T} v_t | \leq \frac{\sqrt{C_1 \beta_T \gamma_T}}{\sqrt{T}} + c + q. \nonumber
\end{align}
where the maximum Gaussian process information gain $\gamma_T$ is bounded above for the squared exponential and Mat\'{e}rn kernels \citep{srinivas2009gaussian}.
\end{thm}

\emph{Proof.} For notational simplicity let us denote the regret term as $\Omega \triangleq \sqrt{C_1 T \beta_T \gamma_T}$. Then from Theorem 1 in \citet{srinivas2009gaussian} we have that
\begin{equation}
\frac{\Omega}{T} \geq f(\lambda^*) - \frac{1}{T} \sum_{t=1}^T f(\lambda_t) \geq f(\lambda^*) - \max_{t \leq T} f(\lambda_t). \label{eq:bound_max}
\end{equation}
This implies $f(\lambda^*) \leq \max_{t \leq T} f(\lambda_t) + \frac{\Omega}{T}$ with probability at least $1 - \frac{\delta}{3}$ (with appropriate choice of $\beta_T$).

Recall that in the proof of Theorem \ref{thm:1} we showed that $|f(\lambda) - f'(\lambda)| \leq c$ 
with probability at least $1 - \frac{\delta}{3}$ (for $c$ given in Algorithm 1). This along with the above expression imply the following two sets of inequalities with probability greater than $1- \frac{2\delta}{3}$:
\begin{align}
\textstyle
f'(\lambda^*) - c \leq f(\lambda^*) <& \; \max_{t \leq T} f(\lambda_t) + \textstyle{\frac{\Omega}{T}}; \nonumber \\
f(\lambda^*) - c \leq f'(\lambda^*) <& \; \max_{t \leq T} f'(\lambda_t) +  \textstyle{\frac{\Omega}{T}}. \nonumber
\end{align}
These, in turn, imply the two sets of inequalities:
\begin{align}
\max_{t \leq T} f'(\lambda_t) \leq f'(\lambda^*) <& \; \max_{t \leq T} f(\lambda_t) + \textstyle{\frac{\Omega}{T}} + c; \nonumber \\
\max_{t \leq T} f(\lambda_t) \leq f(\lambda^*) <& \; \max_{t \leq T} f'(\lambda_t) + \textstyle{\frac{\Omega}{T}} +c. \nonumber
\end{align}
This implies $| \max_{t \leq T} f'(\lambda_t) - \max_{t \leq T} f(\lambda_t) | \leq \frac{\Omega}{T} + c$. That is, the global sensitivity of $\max_{t \leq T} f(\lambda_t)$ is bounded. Given the sensitivity of the maximum $f$, we can readily derive the sensitivity of maximum $v$. First note that we can use the triangle inequality to derive
\begin{align}
| \max_{t \leq T} v'(\lambda_t) - \max_{t \leq T} v(\lambda_t) | \leq& \; | \max_{t \leq T} v(\lambda_t) - \max_{t \leq T} f(\lambda_t) | \nonumber \\
+& \; | \max_{t \leq T} v'(\lambda_t) - \max_{t \leq T} f'(\lambda_t) | \nonumber \\
+& \; | \max_{t \leq T} f'(\lambda_t) - \max_{t \leq T} f(\lambda_t) |. \nonumber
\end{align}
We can immediately bound the final term on the right hand side. Note that as $v_t = f(\lambda_t) + \alpha_t$, the first two terms are bounded above by $|\alpha|$ and $|\alpha'|$, where $\alpha = \{\alpha_{\lceil t \rceil} \mid \lceil t \rceil \triangleq \argmax_{t \leq T} |\alpha_t|\}$ (similarly for $\alpha'$). This is because, in the worst case, the observation noise shifts the observed maximum $\max_{t \leq T} v_t$ up or down by $\alpha$.
Therefore, let $\hat{\alpha} = \alpha$ if $|\alpha| > |\alpha'|$ and $\hat{\alpha} = \alpha'$ otherwise, so that we have:
\begin{equation}
| \max_{t \leq T} v'(\lambda_t) - \max_{t \leq T} v(\lambda_t) | \leq \textstyle{\frac{\Omega}{T}} + c + |2\hat{\alpha}|. \nonumber
\end{equation}

Although $|\hat{\alpha}|$ can be arbitrarily large, recall that for $Z \sim {\cal N}(0,1)$ we have: $\Prob[ |Z| \leq \gamma] \geq 1 - e^{-\gamma^2/2} \triangleq 1 - \frac{\delta}{3}$. Therefore if we set $Z = \frac{2\hat{\alpha}}{\sigma \sqrt{2}}$ we have $\gamma = \sqrt{2 \log(3/\delta)}$. This implies that $|2 \hat{\alpha}| \leq \sigma \sqrt{4 \log(3/\delta)} = q$ with probability at least $1 - \frac{\delta}{3}$. Therefore, if Theorem 1 from \citet{srinivas2009gaussian} and the bound on $| f(\lambda) - f'(\lambda)|$ hold together with probability at least $1- \frac{2\delta}{3}$ as described above, the theorem follows directly. \hfill$\blacksquare$


As in Theorem \ref{lem:sens_x_noise} each quantity in the above bound is given in Algorithm \ref{alg1} ($\beta$, $c$, $q$), given in previous results \citep{srinivas2009gaussian} ($\gamma_T$, $C_1$) or specified by the modeler ($T$, $\delta$). Now that we have a bound on the sensitivity of the maximum $v$ we will use the Laplace mechanism to prove our privacy guarantee (proof in supplementary material):

\begin{coro}
\label{thm:priv_max_f}
Let $\Ac(\Vc)$ denote Algorithm \ref{alg1} run on dataset $\Vc$. Given Assumption \ref{assume:gp}, releasing $\tilde{v}$ is
$(\epsilon,\delta)$-differentially private, i.e., $\Prob[ \Ac(\Vc) \!=\! \tilde{v}] \!\leq\! e^\epsilon \Prob[ \Ac(\Vc') \!=\! \tilde{v} ] \!+\! \delta.$
\end{coro}


Further, as the Laplace distribution has exponential tails, the noise we add to obtain $\tilde{v}$ is not too large:
\begin{thm}
\label{thm:max_f_bound}
Given the assumptions of Theorem \ref{thm:1}, we have the following bound,
\begin{equation}
\textstyle
| \tilde{v} - f(\lambda^*)| \leq \sqrt{2 \log (2T / \delta)} + \frac{\Omega}{T} + a \Big( \frac{\Omega}{\epsilon T} + \frac{c}{\epsilon} + \frac{q}{\epsilon} \Big), \nonumber
\end{equation}
with probability at least $1- (\delta + e^{-a})$ for $\Omega = \sqrt{C_1 T \beta_T \gamma_T}$.
\end{thm}

\emph{Proof (Theorem \ref{thm:max_f_bound}).} Let $Z$ be a Laplace random variable with scale parameter $b$ and location parameter $0$; $Z \sim \mathop{Lap}(b)$. Then $\Prob\bigl[|Z| \leq ab\bigr] = 1 - e^{-a}$. Thus, in Algorithm \ref{alg1}, $| \tilde{v} - \max_{t \leq T} v_t | \leq ab$ for $b = \frac{\Omega}{\epsilon T} + \frac{c}{\epsilon} + \frac{q}{\epsilon}$ with probability at least $1 - e^{-a}$. Further observe,
\begin{align}
\textstyle
ab \geq & \; \max_{t \leq T} v_t - \tilde{v} \geq ( \max_{t \leq T} f(\lambda_t) - \max_{t \leq T} |\alpha_t| )  - \tilde{v} \nonumber \\
\geq & \; f(\lambda^*) - \textstyle{\frac{\Omega}{T}} - \max_{t \leq T} |\alpha_t| - \tilde{v}  \label{eq:third}
\end{align}
where the second and third inequality follow from the proof of Theorem 3 (using the regret bound of \citet{srinivas2009gaussian}: Theorem 1). Note that the third inequality holds with probability greater than $1 - \frac{\delta}{2}$ (given $\beta_t$ in Algorithm \ref{alg1}). The final inequality implies $f(\lambda^*) - \tilde{v} \leq \max_{t \leq T} |\alpha_t| + \frac{\Omega}{T} + ab$. Also note that,
\begin{align}
ab \geq & \; \tilde{v} - \max_{t \leq T} v_t \geq \tilde{v} - (\max_{t \leq T} f(\lambda_t) + \max_{t \leq T} |\alpha_t| ) \nonumber \\
\geq & \;  \tilde{v} - f(\lambda^*) - \textstyle{\frac{\Omega}{T}} - \max_{t \leq T} |\alpha_t|  \label{eq:reverse}
\end{align}
This implies that $f(\lambda^*) - \tilde{v} \geq - \max_{t \leq T} | \alpha_t | - \frac{\Omega}{T} - ab$. Thus we have that $| \tilde{v} - f(\lambda^*)| \leq \max_{t \leq T} |\alpha_t| + \frac{\Omega}{T} + ab$. Finally, because $|\alpha_t|$ could be arbitrarily large we give a high probability upper bound on $|\alpha_t|$ for all $t$. Recall that for $Z_1, \ldots Z_n \sim {\cal N}(0,1)$ we have by the tail probability bound and union bound that $\Prob\bigl[\forall t,\ |Z_t|\!\leq \!\gamma\bigr]\!\geq\! 1-ne^{-\gamma^2/2} \triangleq 1-\frac{\delta}{2}$. Therefore, if we set $Z_t = \alpha_t$ and $n = T$, we obtain $\gamma \!=\! \sqrt{2 \log (2T / \delta)}$. As defined $\gamma \geq \max_{t \leq T} |\alpha_t|$.  \hfill$\blacksquare$

We note that, because releasing either $\tilde{\lambda}$ or $\tilde{v}$ is $(\epsilon,\delta)$-differentially private, by Corollaries \ref{thm:1} and \ref{thm:priv_max_f}, releasing both private quantities in Algorithm 1 guarantees $(2\epsilon,2\delta)$-differential privacy for validation dataset $\Vc$. This is due to the composition properties of $(\epsilon,\delta)$-differential privacy \citep{dwork2006our} (in fact stronger composition results can be demonstrated, \citep{dwork2013algorithmic}).

%% file: exact.tex
In hyper-parameter tuning it may be reasonable to assume that we can observe function evaluations exactly: $v_{\Vc,t} = f_{\Vc}(\lambda_t)$. First note that we can use the same algorithm to report the maximum $\lambda$ in the no-noise setting. Theorems \ref{lem:sens_x_noise} and \ref{thm:2} still hold (note that $q = 0$ in Theorem \ref{thm:2}). However, we cannot readily report a private maximum $f$ as the information gain $\gamma_T$ in Theorems \ref{lem:sense_max_y} and \ref{thm:max_f_bound} approaches infinity as $\sigma^2 \rightarrow 0$. Therefore, we extend results from the previous section to the exact observation case via the regret bounds of \citet{de2012exponential}. Algorithm \ref{alg2} demonstrates how to privatize the maximum $f$ in the exact observation case.

\begin{algorithm}                      
\caption{ Private Bayesian Opt. (noise free obs.)}          
\label{alg2}                           
\begin{algorithmic}                    
	\STATE \textbf{Input:} $\Vc$; $\Lambda \subseteq \mathbb{R}^d$; $T$; $(\epsilon,\delta)$; $A, \tau$; assumptions on $f_\Vc$ in \citet{de2012exponential}	
	\STATE Run method of \citet{de2012exponential}, resulting in noise free observations: $f_{\Vc}(\lambda_1), \ldots, f_{\Vc}(\lambda_T)$
	\STATE $c = 2\sqrt{\bigl(1 - k(\Vc,\Vc')\bigr)\log(2|\Lambda|/\delta)}$
	\STATE Draw $\theta \sim \Lap\!\Big[ \frac{A}{\epsilon} e\!^{- \frac{ 2 \tau }{ (\log 2)^{d/4}}} + \frac{c}{\epsilon}  \Big]$
	\STATE \textbf{Return:} $\tilde{f} = \max_{2 \leq t \leq T} f_{\Vc}(\lambda_t) + \theta$
\end{algorithmic}
\end{algorithm}

\subsection{Private near-maximum validation gain}
\label{sec:f_nonoise}
We demonstrate that releasing $\tilde{f}$ in Algorithm \ref{alg2} is private (Theorem \ref{thm:best_f_noise}) and that a small amount of noise is added to make $\tilde{f}$ private (Theorem \ref{thm:f_utility}). To do so, we derive the global sensitivity of $\max_{2 \leq t \leq T} f(\lambda_t)$ in Algorithm \ref{alg2} independent of the maximum information gain $\gamma_T$ via \citet{de2012exponential}. Then we prove releasing $\tilde{f}$ is $(\epsilon,\delta)$-differentially private and that $\tilde{f}$ is almost $\max_{2 \leq t \leq T} f(\lambda_t)$.

\paragraph{Global sensitivity.}
The following Theorem gives a bound on the global sensitivity of the maximum $f$.

\begin{thm}
\label{lem:f_sensitivity}
Given Assumption \ref{assume:gp} and the assumptions in Theorem 2 of \citet{de2012exponential}, for neighboring datasets $\Vc,\Vc'$ we have the following global sensitivity bound (in the Laplace mechanism sense),
\begin{align}
| \max_{2 \leq t \leq T} f'(\lambda_t) - \max_{2 \leq t \leq T} f(\lambda_t) | \leq A e^{- \frac{ 2 \tau }{ (\log 2)^{d/4}}}+ c \nonumber
\end{align}
w.p.\ at least $1-\delta$ for $c \!=\! 2\sqrt{\bigl(1 \!-\! k(\Vc,\Vc')\bigr)\log(2 |\Lambda|/\delta)}$.
\end{thm}

We leave the proof to the supplementary material. 

Given this sensitivity, we may apply the Laplace mechanism to release $\tilde{f}$.

\begin{coro}
\label{thm:best_f_noise}
Let $\Ac(\Vc)$ denote Algorithm \ref{alg2} run on dataset $\Vc$. Given assumption \ref{assume:gp} and that $f$ satisfies the assumptions of \citet{de2012exponential}, $\tilde{f}$ is $(\epsilon,\delta)$-differentially private, with respect to any neighboring dataset $\Vc'$, \emph{i.e.,}
\begin{align}
\Prob\bigl[ \Ac(\Vc) = \tilde{f}\bigr] \leq e^\epsilon \Prob\bigl[ \Ac(\Vc') = \tilde{f} \bigr] + \delta. \nonumber
\end{align}
\end{coro}

Even though we must add noise to the maximum $f$ we show that $\tilde{f}$ is still close to the optimal $f(\lambda^*)$.

\begin{thm}
\label{thm:f_utility}
Given the assumptions of Theorem \ref{thm:best_f_noise}, we have the utility guarantee for Algorithm \ref{alg2}:
\begin{align}
| \tilde{f} - f(\lambda^*) | \leq \Omega + a \Big( \textstyle{\frac{\Omega}{\epsilon} + \frac{c}{\epsilon}} \Big) \nonumber 
\end{align}
w.p. at least $1 \!-\! (\delta + e^{-a})$ for $\Omega \!=\! Ae^{- \frac{ 2 \tau }{ (\log 2)^{d/4}}}$.
\end{thm}

%% file: no_gp.tex
Even if our our true validation score $f$ is not drawn from a Gaussian process (Assumption \ref{assume:gp}), we can still guarantee differential privacy for releasing its value after Bayesian optimization $f^{\mbox{\scriptsize{BO}}} \!=\! \max_{t \leq T} f(\lambda_t)$. In this section we describe a different functional assumption on $f$ that also yields differentially private Bayesian optimization for the case of machine learning hyper-parameter tuning.



Assume we have a (nonsensitive) training set $\Tc \!=\! \{(\x_i,y_i)\}_{i=1}^n$, which, given a hyperparameter $\lambda$ produces a model $\w(\lambda)$ from the following optimization,
\begin{align}
\w_\lambda = \argmin_{\w} \overbrace{\frac{\lambda}{2} \| \w \|^2_2 + \frac{1}{n} \sum_{i=1}^n \ell(\w, \x_i, y_i)}^{O_{\lambda}(\w)}, \label{eq:opt_train}
\end{align}
The function $\ell$ is a training loss function (e.g., logistic loss, hinge loss). Given a (sensitive) validation set $\Vc = \{(\xbar_i,\ybar_i)\}_{i=1}^m \subseteq \Xc$ we would like to use Bayesian optimization to maximize a validation score $f_{\Vc}$.

\begin{assumption}
\label{assume:smooth_convex} Our true validation score $f_{\Vc}$ is
\begin{align}
f_{\Vc}(\w(\lambda)) = -\frac{1}{m} \sum_{i=1}^m g(\w_\lambda, \xbar_i, \ybar_i), \nonumber
\end{align}
where $g(\cdot)$ is a validation loss function that is $L$-Lipschitz in $\w$ (e.g., ramp loss, normalized sigmoid \citep{huang2014ramp}). Additionally, the training model $\w_\lambda$ is the minimizer of eq.~(\ref{eq:opt_train}) for a training loss $\ell(\cdot)$ that is $1$-Lipschitz in $\w$ and convex (e.g., logistic loss, hinge loss).
\end{assumption}
 

Algorithm \ref{alg3} describes a procedure for privately releasing the best validation accuracy $f^{\BO}$ given assumption \ref{assume:smooth_convex}. Different from previous algorithms, we may run Bayesian optimization in Algorithm \ref{alg3} with any acquisition function (e.g., expected improvement \citep{mockus1978application}, UCB) and privacy is still guaranteed.

\begin{algorithm}                      
\caption{ Private Bayesian Opt. (Lipschitz and convex) }          
\label{alg3}                           
\begin{algorithmic}                    
	\STATE \textbf{Input:} $\Tc$ size $n$; $\Vc$ size $m$; $\Lambda$; $\lambda_{\min}$; $\lambda_{\max}$; $\epsilon$; $T$; $L$; $d$
	\STATE Run Bayesian optimization for $T$ timesteps, observing: $f_{\Vc}(\w_{\lambda_1}), \ldots, f_\Vc(\w_{\lambda_T})$ for $\{ \lambda_1, \ldots, \lambda_T \} = \Lambda_{T,\Vc} \subseteq \Lambda$
	\STATE $f^\BO = \max_{t \leq T} f_\Vc(\w_{\lambda_t})$
	\STATE $g^* = \max_{(\x,y) \in \Xc, \w \in \Rc^d } g(\w,\x,y)$
	\STATE Draw $\theta \sim \Lap\Big[ \frac{1}{\epsilon} \min\{ \frac{g^*}{m}, \frac{L}{m\lambda_{\min} }\} + \frac{(\lambda_{\max} - \lambda_{\min})L }{\epsilon \lambda_{\max} \lambda_{\min}}  \Big]$
	\STATE \textbf{Return:} $\tilde{f}_L = f^\BO + \theta$
\end{algorithmic}
\end{algorithm}

Similar to Algorithms 1 and 2 we use the Laplace mechanism to mask the possible change in validation accuracy when $\Vc$ is swapped with a neighboring validation set $\Vc'$. Different from the work of \citet{chaudhuri2013stability} changing $\Vc$ to $\Vc'$ may also lead to Bayesian optimization searching different hyper-parameters, $\Lambda_{T,\Vc}$ vs. $\Lambda_{T,\Vc'}$. Therefore, we must bound the \emph{total global sensitivity} of $f$ with respect to $\Vc$ \emph{and $\lambda$},

\begin{define}
\label{def:total_global_sense}
The \textbf{total global sensitivity} of $f$ over all neighboring datasets $\Vc,\Vc'$ is 
\begin{align}
\Delta_f \triangleq \max_{ \substack{\Vc,\Vc' \subseteq \Xc \\ \lambda, \lambda' \in \Lambda}} | f_{\Vc}(\w_\lambda) - f_{\Vc'}(\w_{\lambda'}) |.  \nonumber
\end{align}
\end{define}

In the following theorem we demonstrate that we can bound the change in $f$ for arbitrary $\lambda < \lambda'$.

\begin{thm}
\label{thm:f_chg_lam}
Given assumption \ref{assume:smooth_convex}, for neighboring $\Vc,\Vc'$ and arbitrary $\lambda < \lambda'$ we have that,
\begin{align}
\textstyle
| f_{\Vc}(\w_\lambda) \!-\! f_{\Vc'}(\w_{\lambda'}) | \leq \frac{(\lambda' \!-\! \lambda)L}{\lambda' \lambda} + \min \! \big\{ \frac{g^*}{m}, \frac{L}{m\lambda_{\min}} \big\} \nonumber
\end{align}
where $L$ is the Lipschitz constant of $f$, $g^* =  \max_{(\x,y) \in \Xc, \w \in \Rc^d } g(\w,\x,y)$, and $m$ is the size of $\Vc$.
\end{thm}

\emph{Proof.}
Applying the triangle inequality yields
\begin{align}
| f_{\Vc}(\w_\lambda) - f_{\Vc'}(\w_{\lambda'}) | \leq& \; | f_{\Vc}(\w_{\lambda}) - f_{\Vc}(\w_{\lambda'}) | \nonumber \\
+& \; | f_{\Vc}(\w_{\lambda'}) - f_{\Vc'}(\w_{\lambda'}) |. \nonumber
\end{align}
This second term is bounded by \citet{chaudhuri2013stability} in the proof of Theorem 4. The only difference is, as we are not adding random noise to $\w_{\lambda'}$ we have that $| f_{\Vc}(\w_{\lambda'}) - f_{\Vc'}(\w_{\lambda'}) | \leq \min \{ g^*/m, L/(m\lambda_{\min})$ \}.

To bound the first term, let $O_\lambda(\w)$ be the value of the objective in eq.~(\ref{eq:opt_train}) for a particular $\lambda$. Note that $O_\lambda(\w)$ and $O_{\lambda'}(\w)$ are $\lambda$ and $\lambda'$-strongly convex. Define
\begin{align}
\textstyle
h(\w) = O_{\lambda'}(\w) - O_{\lambda}(\w) = \frac{\lambda' - \lambda}{2} \| \w \|^2_2.
\end{align}
Further, define the minimizers $\w_\lambda \!=\! \argmin_\w O_{\lambda}(\w)$ and ${\w}_{\lambda'} \!=\! \argmin_\w [O_{\lambda}(\w) + h(\w)]$. This implies that
\begin{align}
\nabla O_{\lambda}(\w_\lambda) = \nabla O_{\lambda}({\w}_{\lambda'}) + \nabla h({\w}_{\lambda}) = 0. \label{eq:grad}
\end{align}
Given that $O_\lambda$ is $\lambda$-strongly convex \citep{shalev2007online}, and by the Cauchy-Schwartz inequality,
\begin{align}
\lambda & \; \|\w_{\lambda} \!-\! \w_{\lambda'}\|^2_2 \leq \Big[\nabla O_\lambda(\w_\lambda) - \nabla O_{\lambda}({\w}_{\lambda'})\Big]^\top\Big[\w_\lambda \!-\! {\w}_{\lambda'}\Big] \nonumber \\
\leq& \; \nabla h(\w_{\lambda'})^\top\Big[\w_{\lambda} \!-\! \w_{\lambda'}\Big] \leq \|\nabla h(\w_{\lambda'}) \|_2 \|\w_{\lambda} \!-\! \w_{\lambda'}\|_2. \nonumber
\end{align}
Rearranging, 
\begin{align}
\frac{1}{\lambda} \|\nabla h(\w_{\lambda'}) \|_2 = \Big\| \frac{\lambda' \!-\! \lambda}{2} \nabla \|\w_{\lambda'}\|_2^2 \Big\|_2 \geq \| \w_{\lambda} \!-\! \w_{\lambda'} \|_2 \label{eq:what_h_is}
\end{align}
Now as $\w_{\lambda'}$ is the minimizer of $O_{\lambda'}$ we have, 
\begin{align}
\textstyle
\nabla \| \w_{\lambda'} \|_2^2 = \frac{2}{\lambda'} \big[ - \frac{1}{n} \sum_{i=1}^n \nabla \ell(\w_{\lambda'}, \x_i, y_i) \big]. \nonumber
\end{align}
Substituting this value of $\w_{\lambda'}$ into eq.~(\ref{eq:what_h_is}) and noting that we can pull the positive constant term $(\lambda' - \lambda)/2$ out of the norm and drop the negative sign in the norm gives us
\begin{align}
\textstyle
\frac{1}{\lambda} \|\nabla h(\w_{\lambda'}) \|_2 \!=\! \frac{\lambda' \!-\! \lambda}{\lambda \lambda'} \Big\| \frac{1}{n} \sum_{i=1}^n \nabla \ell(\w_{\lambda'}, \x_i, y_i)  \Big\|_2 \!=\! \frac{\lambda' \!-\! \lambda}{\lambda \lambda'}. \nonumber
\end{align}
The last equality follows from the fact that the loss $\ell$ is $1$-Lipschitz by Assumption \ref{assume:smooth_convex} and the triangle inequality.
Thus, along with eq.~(\ref{eq:what_h_is}), we have
\begin{align}
\textstyle
\| \w_{\lambda} \!-\! \w_{\lambda'} \|_2 \leq \frac{1}{\lambda} \|\nabla h(\w_{\lambda'}) \|_2 \leq \frac{\lambda' - \lambda}{\lambda \lambda'}. \nonumber 
\end{align}
Finally, as $f$ is $L$-Lipschitz in $\w$,
\begin{align}
\textstyle
|f_{\Vc}(\w_\lambda) - f_{\Vc}(\w_{\lambda'})| \leq L \| \w_{\lambda} - \w_{\lambda'} \|_2 \leq  L \frac{\lambda' - \lambda}{\lambda \lambda'} \nonumber 
\end{align}
Combining the result of \citet{chaudhuri2013stability} with the above expression completes the proof.
\hfill$\blacksquare$

Given a finite set of possible hyperparameters $\Lambda$, we would like to bound $| f_\Vc^* - f_{\Vc'}^* |$; the best validation score found when running Bayesian optimization on $\Vc$ vs. $\Vc'$. Note that, by Theorem 7,
\begin{align}
\textstyle
| f_\Vc^* - f_{\Vc'}^* | \leq & \; \max_{\lambda,\lambda'} | f_{\Vc}(\w_\lambda) \!-\! f_{\Vc'}(\w_{\lambda'}) | \nonumber \\
\leq & \; \textstyle{\frac{(\lambda_{\max} \!-\! \lambda_{\min})L}{\lambda_{\max} \lambda_{\min}} + \min \! \Big\{ \frac{g^*}{m}, \frac{L}{m\lambda_{\min}}} \Big\}, \nonumber
\end{align}
as $(\lambda' \!-\! \lambda)/(\lambda' \lambda)$ is strictly increasing in $\lambda'$ strictly decreasing in $\lambda$. Given this sensitivity of $f^*$ we can use the Laplace mechanism to hide changes in the validation set as follows. 

\begin{coro}
Let $\Ac(\Vc)$ denote Algorithm \ref{alg3} applied on dataset $\Vc$. Given assumption \ref{assume:smooth_convex}, $\tilde{f}_L$ is $\epsilon$-differentially private, i.e., 
$\Prob\bigl[ \Ac(\Vc) = \tilde{f}_L\bigr] \leq e^\epsilon \Prob\bigl[ \Ac(\Vc') = \tilde{f}_L\bigr]$ 
\end{coro}
We leave the proof to the supplementary material. Further, by the exponential tails of the Laplace mechanism we have the following utility guarantee,
\begin{thm}
Given the assumptions of Theorem \ref{thm:f_chg_lam}, we have the following utility guarantee for $\tilde{f}_L$ w.r.t. $f^\BO$,
\begin{align}
\textstyle
| \tilde{f}_L - f^\BO | \leq a\Big[ \frac{1}{\epsilon m} \min\{ g^*, \frac{L}{\lambda_{\min}}  \} + \frac{(\lambda_{\max} - \lambda_{\min}) L }{ \epsilon \lambda_{\max} \lambda_{\min} } \Big] \nonumber
\end{align}
with probability at least $1 - e^{-a}$.
\end{thm}

\emph{Proof.} This follows exactly from the tail bound on Laplace random variables, given in the beginning of the proof of Theorem 6. \hfill$\blacksquare$


%% file: results.tex
In this section we examine the validity of our multi-task Gaussian process assumption on $[f_1, \ldots, f_{2^{|\Xc|}}]$. Specifically, we search for the most likely value of the multi-task Gaussian process covariance element $k_1(\Vc,\Vc')$, for classifier hyper-parameter tuning. Larger values of $k_1(\Vc,\Vc')$ correspond to a smaller global sensitivity bounds in Theorems \ref{lem:sens_x_noise}, \ref{lem:sense_max_y}, and \ref{lem:f_sensitivity} leading to improved privacy guarantees.

For our setting of hyper-parameter tuning, each $\lambda = [C, \gamma^2]$ are hyper-parameters for training a kernelized support vector machine (SVM) \citep{cortes1995support,scholkopf2001learning} with cost parameter $C$ and radial basis kernel width $\gamma^2$. The value $f_{\Vc}(\lambda)$ is the accuracy of the SVM model trained with hyper-parameters $\lambda$ on $\Vc$.

\begin{figure}[t]
\includegraphics[width=\columnwidth]{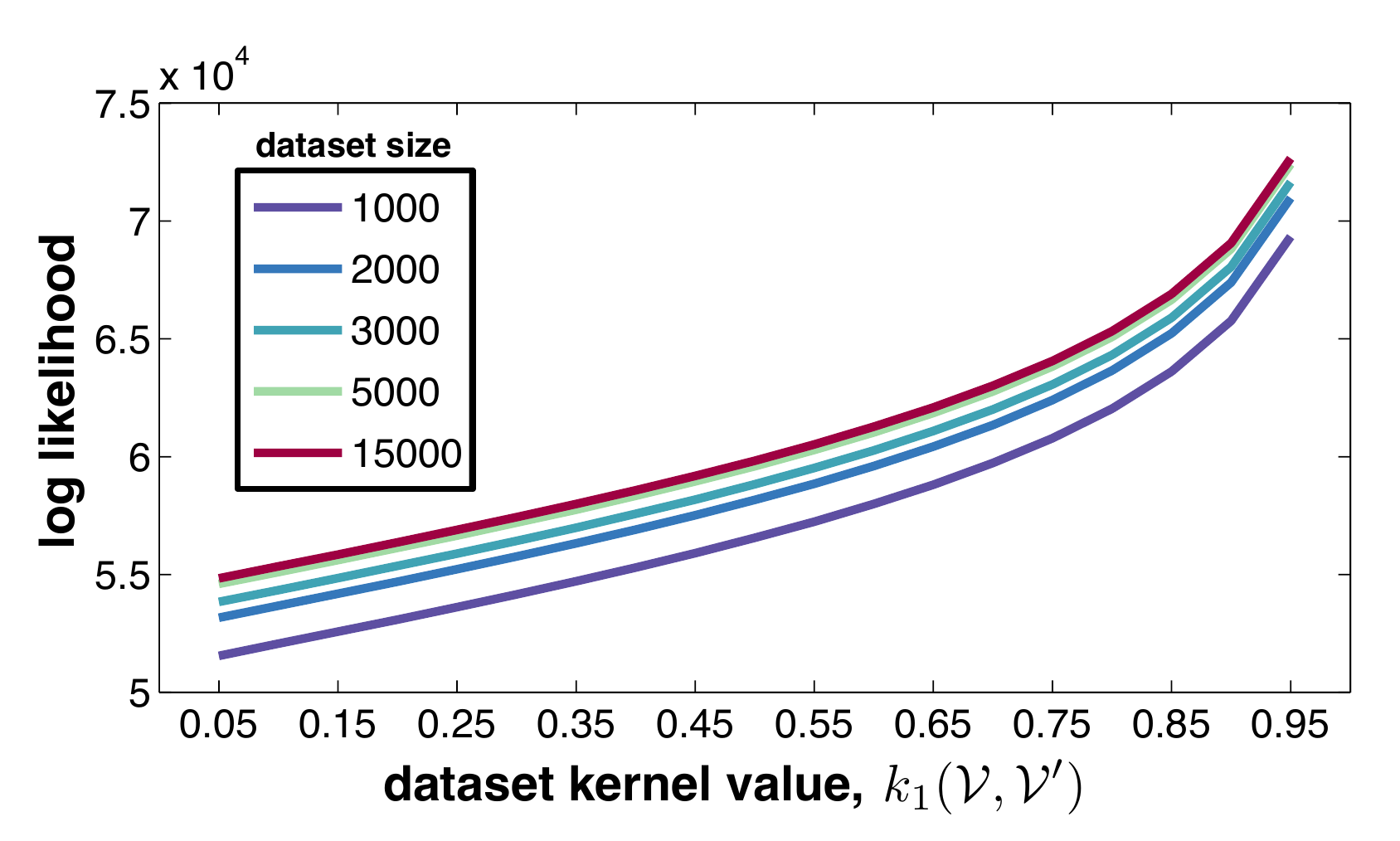}
\vskip -0.2in
\caption{The log likelihood of a multi-task Gaussian process for different values of the kernel value $k_1(\Vc,\Vc')$. The function evaluations are the validation accuracy of SVMs with different hyper-parameters.}
\label{fig:log_like}
\end{figure}

To search for the most likely $k_1(\Vc,\Vc')$ we start by sampling $100$ different SVM hyper-parameter settings $\lambda_1,\ldots,\lambda_{100}$ from a Sobol sequence 
and train an SVM model for each on the Forest UCI dataset ($36,603$ training inputs). We then randomly sample $100$ i.i.d.\ validation sets $\Vc$. Here we describe the evaluation procedure for a fixed validation set size, which corresponds to a single curve in Figure \ref{fig:log_like} (as such, to generate all results we repeat this procedure for each validation set size in the set $\{1000,2000,3000,5000,15000\}$).
 For each of the $100$ validation sets we randomly add or remove an input to form a neighboring dataset $\Vc'$. We then evaluate each of the trained SVM models on all $100$ datasets $\Vc$ and their pairs $\Vc'$. This results in two $100 \times 100$ (number of datasets, number of trained SVM models) function evaluation matrices $\mathbf{F}_{\Vc}$ and $\mathbf{F}_{\Vc'}$. Thus, $[\mathbf{F}_{\Vc}]_{ij}$ is the validation accuracy on the $i^{\scriptsize \mbox{th}}$ validation set $\Vc_i$ using the $j^{\scriptsize \mbox{th}}$ SVM model.

The likelihood of function evaluations for a dataset pair $(\Vc_i,\Vc_i')$, for a value of $k_1(\Vc_i,\Vc_i')$, is given by the marginal likelihood of the multi-task Gaussian process:
\begin{align}
\Prob
\begin{pmatrix}
[\mathbf{F}_{\Vc}]_i \\
[\mathbf{F}_{\Vc'}]_i \\
\end{pmatrix}
\sim {\cal N}( \mu_{100}, \sigma_{100}^2 ), \label{eq:marg_like}
\end{align}
where $[\mathbf{F}_{\Vc}]_i = [f_{\Vc}(\lambda_1), \ldots, f_{\Vc}(\lambda_{100})]$ (similarly for $\Vc'$) and $\mu_{100}$ and $\sigma_{100}^2$ are the posterior mean and variance of the multi-task Gaussian process using kernel $k_1(\Vc,\Vc') \otimes k_2(\lambda,\lambda')$ after observing $[\mathbf{F}_{\Vc}]_i$ and $[\mathbf{F}_{\Vc'}]_i$ (for more details see \citet{bonilla2008multi}). As $\mu_{100}$ and $\sigma_{100}^2$ depend on $k_1(\Vc,\Vc')$, we treat it as a free-parameter and vary its value from $0.05$ to $0.95$ in increments of $0.05$. For each value, we compute the marginal likelihood (\ref{eq:marg_like}) for all validation datasets ($\Vc_i$ for $i = 1,\ldots,100$). As each $\Vc_i$ is sampled i.i.d.\, the joint marginal likelihood is simply the product of all $\Vc_i$ likelihoods. Computing this joint marginal likelihood for each $k_1(\Vc,\Vc')$ value yields a single curve of Figure \ref{fig:log_like}. As shown, the largest values of $k_1(\Vc,\Vc') = 0.95$ is most likely, meaning that $c$ in the global sensitivity bounds is quite small, leading to private values that are closer to their true optimums.




%% file: related.tex
There has been much work towards differentially private convex optimization \citep{chaudhuri2011differentially, kifer2012private, duchi2013local, song2013stochastic, jain2014near, bassily2014private}. The work of \citet{bassily2014private} established upper and lower bounds for the excess empirical risk of $\epsilon$ and $(\epsilon,\delta)$-differentially private algorithms for many settings including convex and strongly convex risk functions that may or may not be smooth. There is also related work towards private high-dimensional regression, where the dimensions outnumber the number of instances \citep{kifer2012private,thakurta2013differentially}. In such cases the Hessian becomes singular and so the loss is nonconvex. However, it is possible to use the restricted strong convexity of the loss in the regression case to guarantee privacy. 

Differential privacy has been shown to be achievable in online and interactive kernel learning settings \citep{jain2011differentially,thakurta2013nearly,jain2013differentially,mishraprivate}. In general, non-private online algorithms are closest in spirit to the methods of Bayesian optimization. However, all of the previous work in differentially private online learning represents a dataset as a sequence of bandit arm pulls (the equivalent notion in Bayesian optimization is function evaluations $f(\x_t)$). Instead, we consider functions in which changing a single dataset entry possibly affects \emph{all future function evaluations}. Closest to our work is that of \citet{chaudhuri2013stability}, who show that given a fixed set of hyper-parameters which are always evaluated for any validation set, they can return a private version of the index of the best hyper-parameter, as well as a private model trained with that hyper-parameter. Our setting is strictly more general in that, if the validation set changes, Bayesian optimization could search completely different hyper-parameters.

Bayesian optimization, largely due to its principled handling of the exploration/exploitation trade-off of global, black-box function optimization, is quickly becoming the global optimization paradigm of choice. Alongside promising empirical results there is a wealth of recent work on convergence guarantees for Bayesian optimization, similar to those used in this work \citep{srinivas2009gaussian,de2012exponential}. \citet{vazquez2010convergence} and \citet{bull2011convergence} give regret bounds for optimizing the expected improvement acquisition function each optimization step. BayesGap \citep{hoffman2014correlation} gives a convergence guarantee for Bayesian optimization with budget constraints. Bayesian optimization has also been extended to multi-task optimization \citep{BaBrKeSe13,swersky2013multi}, the setting where multiple experiments can be run at once \citep{azimi2012hybrid,snoek2012practical}, and to constrained optimization \citep{gardner2014bayesian}.


%% file: conclusion.tex
We have introduced methods for privately releasing the best hyper-parameters and validation accuracies in the case of exact and noisy observations. Our work makes use of the differential privacy framework, which has become commonplace in private machine learning \citep{dwork2013algorithmic}. We believe we are the first to demonstrate differentially private quantities in the setting of global optimization of expensive (possibly nonconvex) functions, through the lens of Bayesian optimization.

One key future direction is to design techniques to release each sampled hyper-parameter and validation accuracy privately (during the run of Bayesian optimization). This requires analyzing how the maximum upper-confidence bound changes as the validation dataset changes. Another interesting direction is extending our guarantees in Sections \ref{sec:noise} and \ref{sec:exact} to other acquisition functions.

For the case of machine learning hyper-parameter tuning our results are designed to guarantee privacy of the validation set only (it is equivalent to guarantee that the training set is never allowed to change). To simultaneously protect the privacy of the training set it may be possible to use techniques similar to the training stability results of \citet{chaudhuri2013stability}. Training stability could be guaranteed, for example, by assuming an additional training set kernel that bounds the effect of altering the training set on $f$. We leave developing these guarantees for future work.

As practitioners begin to use Bayesian optimization in practical settings involving sensitive data, it suddenly becomes crucial to consider how to preserve data privacy while reporting accurate Bayesian optimization results. This work presents methods to achieve such privacy, which we hope will be useful to practitioners and theorists alike.